\documentclass[preprint,12pt]{elsarticle}

\usepackage{amssymb}

\begin{document}

\begin{frontmatter}



\title{Object Recognition System on a Tactile Device for Visually Impaired}


\author{Souayah Abdelkader, MOKRETAR KRAROUBI Abderrahmene, Slimane LARABI}

\affiliation{organization={USTHB University},
            addressline={BP 32 El Alia}, 
            city={Algiers},
            postcode={16111}, 
            country={Algeria}}

\begin{abstract}
People with visual impairments face numerous challenges when interacting with their environment. Our objective is to develop a device that facilitates communication between individuals with visual impairments and their surroundings. The device will convert visual information into auditory feedback, enabling users to understand their environment in a way that suits their sensory needs.
Initially, an object detection model is selected from existing machine learning models based on its accuracy and cost considerations, including time and power consumption. The chosen model is then implemented on a Raspberry Pi, which is connected to a specifically designed tactile device. When the device is touched at a specific position, it provides an audio signal that communicates the identification of the object present in the scene at that corresponding position to the visually impaired individual.
Conducted tests have demonstrated the effectiveness of this device in scene understanding, encompassing static or dynamic objects, as well as screen contents such as TVs, computers, and mobile phones.

\end{abstract}







\end{frontmatter}









\section{Introduction}

People with visual impairments face numerous challenges in their daily lives. They are unable to perceive the world in the same way as those with sight and encounter multiple difficulties, including orientation, obstacle detection and avoidance, limited mobility, and an inability to recognize shapes and colors of objects in their surroundings. In addition to these challenges, they are completely excluded from understanding and interacting with the real world scene.

Numerous technological advancements have been made to assist people with visual impairments. Among the different technological solutions deployed to address this specific need, computer vision-based solutions appear as one of the most promising options due to their affordability and accessibility.

Systems with human-scene interaction generate outputs after processing the captured scene. They consist of a set of computer vision and machine learning techniques aimed at improving the user's life in various activities such as content interpretation, navigation, etc. Generally, these systems process the data received from the real world using depth or RGB sensors and transform them into instructions and signals ~\cite{these-chaymal-arabi} ~\cite{CCSSP2020}.

\begin{figure}[t]
\centering
\includegraphics[width=10cm]{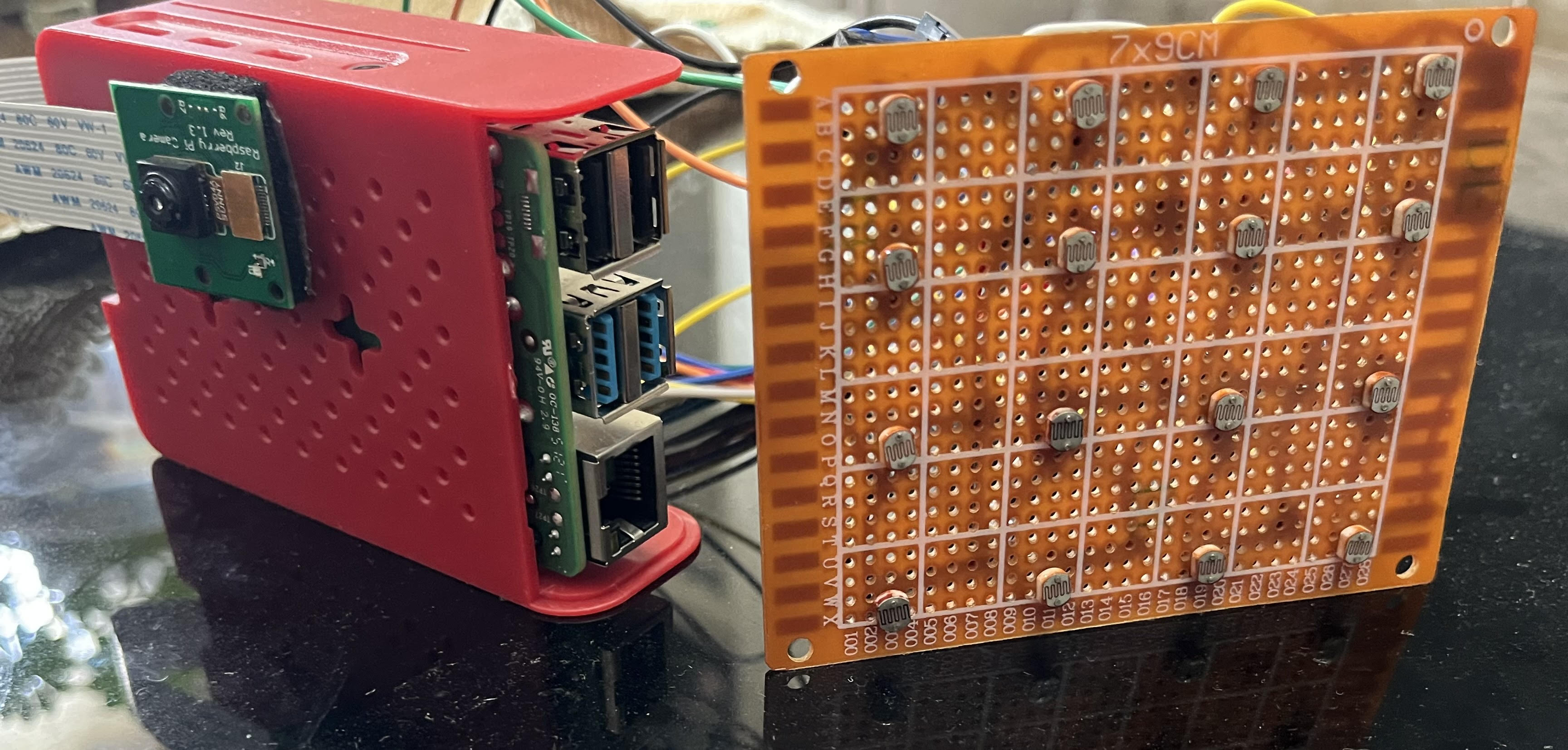}
\caption{The prototype of the proposed system.}
\label{fig:fig1}
\end{figure}

The goal of this work is to assist individuals with visual impairments in perceiving the information contained in an image by displaying the coded scene on a tactile device. They can explore the image by touching the pins on the device, with each pin representing a corresponding object in the scene.
The developed system prototype is illustrated in Figure \ref{fig:fig1}.

This paper is organized as follows. In the next section, we review the main deep learning methods for object detection and compare these methods to select the most accurate model in terms of time and precision. 
The third section is devoted to proposed system and includes the design of the device, which takes the identities of recognized objects as input and outputs a signal for human-machine interaction.
In Section 4, we present the implementation of the system on a Raspberry Pi board and the conducted tests. The results are presented and discussed.
We conclude with some future perspectives.

\section{Related Works}

Object detection has been well studied and recent models give accurate precision. In  ~\cite{Diwan2023} ~\cite{naftali2022} two major categories of object detection networks have been observed ~\cite{Arkin2022}:\\
Two-stage networks, were historically the first ones used for object detection ~\cite{Jiao_2019}. They employ two successive neural networks, called stages: the first one is a region proposal network (Region Proposal Network), which proposes potential bounding boxes. The second stage regresses the position and label of the bounding boxes. Networks in this category include R-CNN ~\cite{girshick2014rich}, which was later improved by Fast R-CNN ~\cite{girshick2015fast} and Faster R-CNN ~\cite{ren2016faster}. These networks perform better in terms of precision metric but at the expense of inference speed metric.\\
One-stage networks, also known as one-stage, have only one stage responsible for generating bounding boxes and labels. Networks in this category include RetinaNet \cite{8417976}, SSD (Single-Shot Multibox Detector) \cite{Liu_2016}, and the YOLO (You Only Look Once) family of networks \cite{wang2021scaledyolov4}. These networks are less accurate in terms of precision metric but are faster.\\
Object detection models based on Transformers use neural network architectures that rely on attention mechanisms, enabling them to consider different parts of the image at various levels of abstraction \cite{dosovitskiy2021image}.
These methods have recently demonstrated impressive performance on object detection tasks, with results comparable to or surpassing those of traditional object detection methods based on CNNs \cite{Arkin2022}.

Once objects are detected, they serve to build systems to assist visually impaired individuals in various ways including understanding the scene from depth images ~\cite{these-chaymal-arabi} ~\cite{Zatout2019} ~\cite{CCSSP2020}, identifying objects ~\cite{34}, navigating their environment and avoiding obstacles ~\cite{31}, visual positioning from depth images ~\cite{Ibel2022} ~\cite{Ibel2020} ~\cite{Ibel2020_2}, image captioning for visually impaired ~\cite{Delloul2022_2} ~\cite{Delloul2022} and  Human Action Recognition and Coding based on Skeleton ~\cite{Benhamida2022} ~\cite{benhamida2023theater}. 

 In general, human-scene interaction aims to facilitate understanding and exploration of the environment for visually impaired individuals, and it can be categorized as Tactile-Sound Interaction and Tactile-Tactile Interaction.
In \cite{these-chaymal-arabi}, the authors proposed a method for semantic scene labeling using RGB-D images to facilitate human-scene interaction. The obtained objects are converted into semantic codes inspired by the Braille system and the Japanese Kanji writing system. 

Additionally, work has been done based on the sonification of images. In \cite{CAVACO2013}, a software tool was developed to assist visually impaired individuals in identifying the color and brightness of an image through sonification. This software tool extracts color information from an image or video using HSV (hue, saturation, value) information, which is then converted into audio attributes such as pitch, timbre, and loudness. This tool can be used to gather information about the range of colors present in images, the presence or absence of light sources, as well as the location and shape of objects in the images.

image sonification was also developed has been proposed n ~\cite{Banf2013}~\cite{Banf2016}, where individuals would actively explore an image on a touchscreen and receive auditory feedback on the content of the image at the current position. In this system, feature extracted  and classified, objects are detected and recognized and 
are acoustically represented using drum sounds.

Even if such solutions can help individuals to understand the scene content, in majority of cases, they do not need to explore the details of the image content but the content of the scene in terms of objects and in dynamic case, image exploration became more and more heavy.
In this direction, we propose to use the latest technologies of deep learning to build a system that help individual to be informed about the surrounding scenes even if is moving. 

\section{Proposed System}

Our system aims to assist visually impaired individuals in identifying objects and their locations from images. A tactile device has been developed to provide auditory feedback corresponding to the identity of the detected object, thereby helping these individuals obtain information about the scene. The proposed system is capable of identifying 17 types of objects in the observed scene.

\begin{figure}[t]
\begin{center}
   \includegraphics[width=12cm]{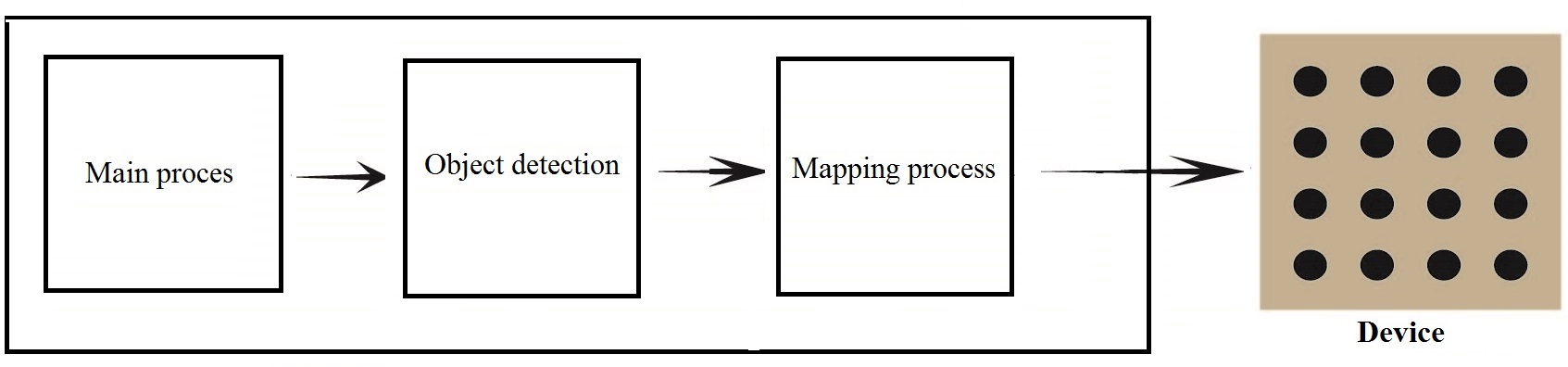}
\end{center}
   \caption{The proposed system}
   \label{fig:schema}
\end{figure}

This system is divided into three processes as illustrated in Figure~\ref{fig:schema}.

The first two processes cooperate in interpreting and detecting objects in the observed scene. Since the system is embedded on a Raspberry Pi, it has limited resources (low RAM and processing power). Therefore, we have developed three object detection models, each responsible for detecting objects in a specific environment: Office, Kitchen, and Bedroom.

The main process is responsible for recognizing the appropriate environment in order to load the corresponding model. This process is reactivated at each new camera location and when the detection rate falls below a predefined threshold.

The second process is responsible for detecting objects and their locations in the image, and it transfers the coordinates of the object locations to the next process.

The third process involves associating the detected objects with a location on the tactile device and interacting with the user to produce corresponding sound feedback for the detected object.

\subsection{The Detection Processes}

The acquired image is input into the main process, which determines the appropriate environment or scene category (e.g., office, kitchen, bedroom) based on the visual cues and characteristics present in the image.
These three models are based on YOLOv5 and have been retrained on a dataset consisting of seven specific object classes. The goal of each model is to detect and recognize objects belonging to these seven classes in their corresponding environment.

The system operates using an object detection model that is responsible for detecting characteristic objects in each environment. Then, the k-nearest neighbors algorithm is executed to recognize the observed environment. In this algorithm, objects represent the features, and environments represent the target classes.

Once the appropriate environment is determined, the second process takes over for object detection and location, while the first process is paused. It transfers the coordinates of each detected object to the final process. This process also has the responsibility of reactivating the main process when the detection rate falls below a predefined threshold (e.g., when the object detection model fails to detect more than 20\% of the objects in the observed environment).

\subsection{Mapping process}

This process is responsible for converting the coordinates of objects in the image into relative coordinates on the tactile device. In cases where there is overlap between two objects, where both objects may appear in the same grid cell of the tactile device, or when a relatively large object occupies multiple grid cells, we have developed an algorithm to determine the order of objects belonging to the same grid cell.

Additionally, this process is responsible for interacting with the user through the tactile device. It produces sound feedback corresponding to each detected object. When the user touches or interacts with a specific pin on the tactile device, a specific sound is emitted to provide feedback to the user.

\subsection{Model selection}

In order to select the appropriate model for the specific task of integrating an object detection model into a Raspberry Pi, we conducted a comparative study of object detection algorithms based on Convolutional Neural Networks (CNNs).

Considering our objective of achieving acceptable precision and recall values while working with embedded systems, we conducted the comparison while considering the following constraints:

We focused on using the latest reduced versions of each model, commonly referred to as "tiny models," such as YOLOv5, Faster R-CNN, and SSD.
The object detection models used in the comparison were trained on the same image dataset and shared the same backbone architecture. This ensured that we could make meaningful observations regarding the advantages and disadvantages of these methods.

Initially, we conducted a comparison using the publicly available MS-COCO dataset ~\cite{Lin2014}. We selected 20 classes with over 200 images per class. These datasets include classes of different sizes, ranging from small classes like spoons, mice, and remote controls, to medium-sized classes like televisions and laptops, and large classes like people, beds, and dining tables. These selected classes represent three environments: office, kitchen, and bedroom, to ensure a relatively accurate comparison. Additionally, we compared these classes using hundreds of images collected from the internet and captured with a Raspberry Pi device.

Initially, we conducted a comparison using the publicly available MS-COCO dataset ~\cite{Lin2014}. We selected 20 classes with over 200 images per class from this dataset. These selected classes encompass a range of sizes, including small classes like spoons, mice, and remote controls, medium-sized classes like televisions and laptops, and large classes like people, beds, and dining tables. By including these diverse classes, we aimed to ensure a comprehensive and representative comparison.

To further enhance the accuracy of the comparison, we also incorporated additional datasets consisting of hundreds of images collected from the internet and captured with a Raspberry Pi device. These images were carefully chosen to cover the selected classes and represent three specific environments: office, kitchen, and bedroom. By including images captured with the Raspberry Pi device, we aimed to account for any specific characteristics or challenges that may arise when using the object detection models in an embedded system setup.

Tables \ref{tab:resultat1} and \ref{tab:resultat2} present the results obtained on the MS-COCO benchmark and a collected image dataset in terms of mAP0.5 (mean Average Precision at IoU threshold of 0.5). These results validate the effectiveness of YOLOv5 in comparison to Faster R-CNN and SSD. The tables demonstrate that the YOLOv5 structure is better suited for real-time applications due to its faster processing speed compared to the other structures.

\begin{table}
    \begin{center}
    \scalebox{0.8}{
   \begin{tabular}{|l|c|c|c|c|c|}
        \hline
          & Image dimension &	Dataset &	Backbone &	Inference time &	mAP.5 \\
        \hline 
        Yolov5 &	640*426 &	MS COCO &	Tiny version of  & 	9 ms &	0.53 \\
         &  &  &  CSP-Darknet53 &  &  \\
         \hline
        Fasterrcnn & 	640*426 &	MS COCO &	ResNet-50 &
	68.54 ms &	0.49 \\
        \hline
        ssd &	640*426 &	MS COCO &	ResNet-50 &
	12.6 ms &	0.21  \\
        \hline 
\end{tabular} }
\end{center}
    \caption{Comparison results on public database.}
    \label{tab:resultat1}
\end{table}  

 \begin{table}
    \centering
    \scalebox{0.9}{
   \begin{tabular}{|l|c||c||c|}
        \hline
          &	Backbone &	Inference time &	mAP.5 \\
        \hline
        Yolov5 &		Tiny version of CSP-Darknet53 & 	10.18 ms &	0.64 \\
        \hline
        Fasterrcnn & 	ResNet-50 &	92.67 ms &	0.63 \\ 
        \hline
        ssd &	ResNet-50 &	15.84 ms &	0.40 \\
        \hline
\end{tabular} }
    \caption{Comparison results on our collected images.}
    \label{tab:resultat2}
\end{table}  

The selected model, determined by the main process, utilizes a YOLOv5-based object detection method to identify and locate objects in the image. It generates bounding boxes that enclose each detected object, accompanied by confidence scores that must exceed 0.5 to be deemed valid.

The coordinates of the detected objects, represented by the bounding boxes, are extracted from the object detection model and transmitted to the final process for encoding them on the tactile device.

\begin{figure}[t]
    \centering
    \includegraphics[width=12cm]{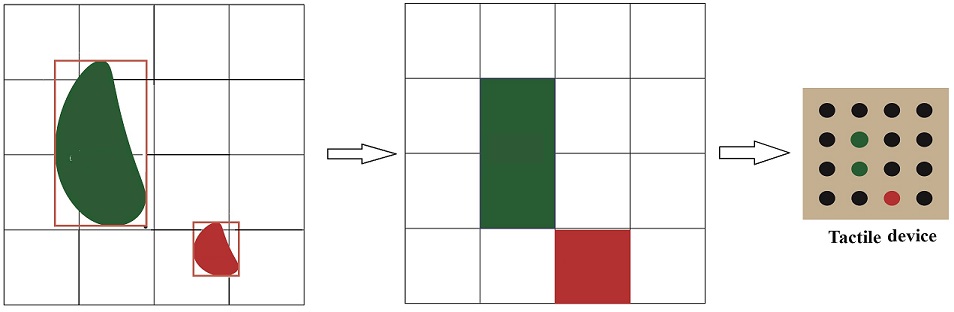}
    \caption{Mapping objects into cells on the device.}
    \label{fig:mapping.}
\end{figure}

\section{Experimental Results}

\begin{figure}[t]
   \centering
  \includegraphics[width=14cm]{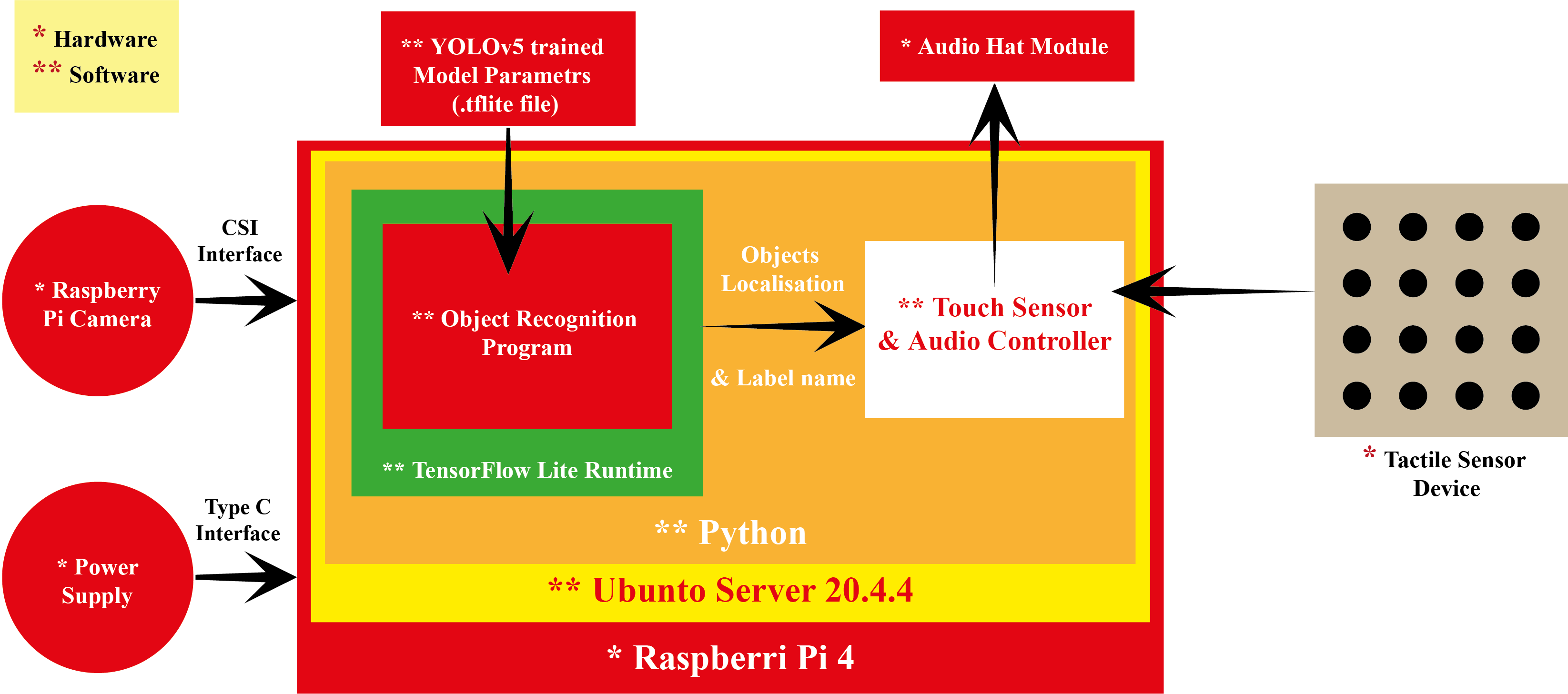}
    \caption{Components of the system}
   \label{fig:fig2.}
\end{figure}

In Figure ~\ref{fig:fig2.}, we present the components of our interactive device designed for visually impaired individuals.
This compact and portable device is a Raspberry Pi equipped with a high-definition camera, a 2GB CPU, and RAM. The device analyzes the user's environment and detects objects in real-time. The gathered information is then transmitted to the user through a haptic feedback system.
The haptic feedback system utilizes a device with 16 photoresistor sensors, enabling visually impaired individuals to comprehend their environment using their fingers. These sensors detect the presence of fingers and convert this information into audio feedback.

\subsection{Making the device}

Our main goal is to enable tactile-audio interaction with visually impaired users. To accomplish this, we have opted to utilize photoresistor technology, an electronic component that exhibits varying electrical resistance in response to incident light. The resistance of a photoresistor changes inversely proportional to the intensity of light it receives. In tactile interaction, this technology can be employed to detect changes in light caused by the user's touch on a light-sensitive surface.

By arranging multiple photoresistors as pins on a surface, we can detect which resistors are touched by the user. This enables tactile interaction where the user can interact with different pins and trigger actions, such as audio feedback through an audio output module connected to the Raspberry Pi. Each touched resistor corresponds to a specific sound based on the detected object's position in the image relative to the pin.

\begin{figure}[t]
\centering
 \includegraphics[width=10cm]{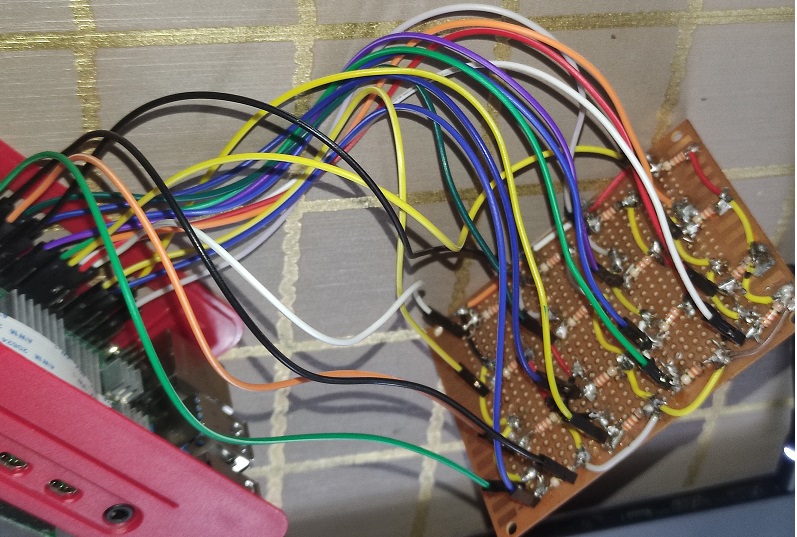}
\caption{Connecting Raspberry Pi and the tactile device.}
\label{fig:connexion}
\end{figure}

Figure~\ref{fig:connexion} illustrates the circuit connecting 16 photoresistors, with each photoresistor connected to one of the Raspberry Pi's pins. The associated code utilizes the RPi.GPIO library to manage GPIO pins on the Raspberry Pi. It configures the port for the photoresistor as an input. In the main loop, it checks the state of the photoresistor. If it is triggered (HIGH), it displays a message indicating that the user has touched the photoresistor.

\subsection{Object Detection}
\begin{figure}[t]
 \centering
 \includegraphics[width=5cm]{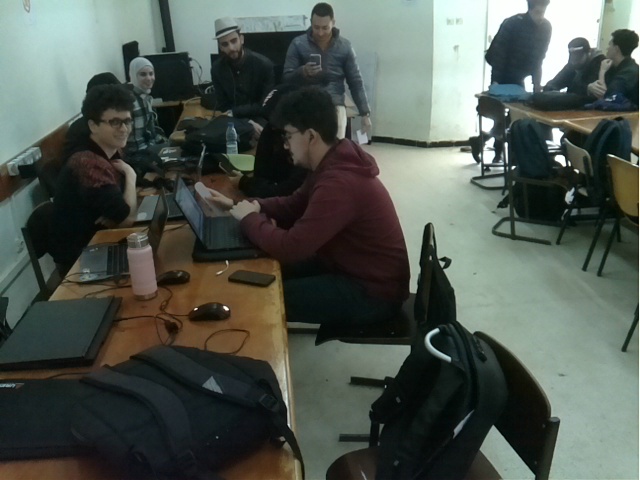}
 \includegraphics[width=5cm]{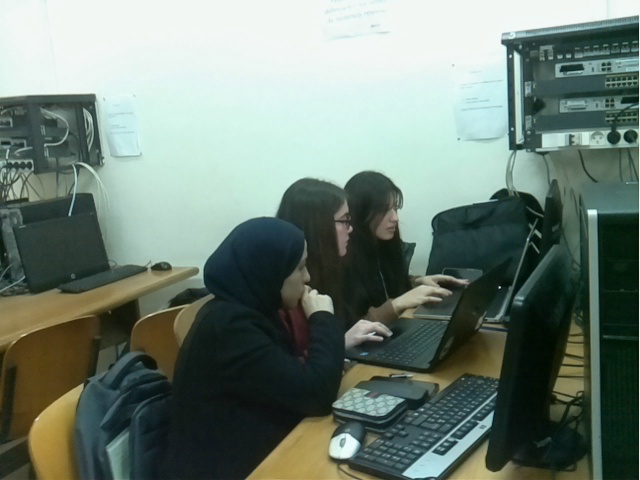}\\
 \includegraphics[width=5cm]{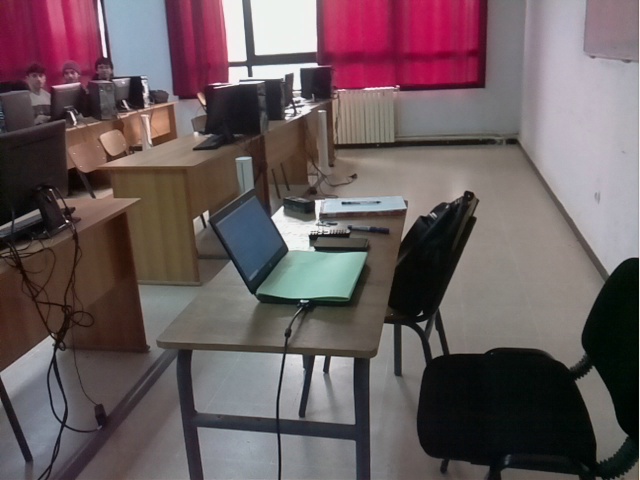}
  \includegraphics[width=5cm]{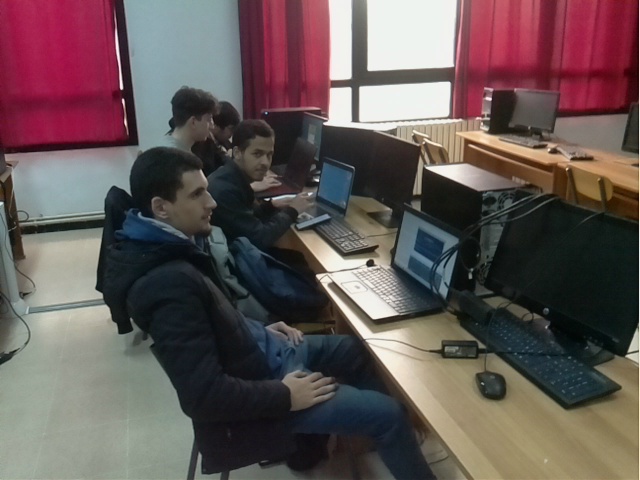}
 \caption{Some acquired images.}
\label{fig:ensemble_images}
\end{figure}

YOLOv5 is an extremely efficient object detection algorithm based on deep learning. It has been trained on a dataset that includes images acquired at our faculty (refer to Fig~\ref{fig:ensemble_images}), as well as a collection of publicly available images. The training process is conducted using PyTorch and the TensorFlow Lite platform.

By utilizing connected modules on the tactile device, the video images captured by the piCamera module are analyzed to search for objects that have been learned by the embedded model on the Raspberry Pi. Subsequently, a corresponding sound associated with each detected object is played.

The primary objective of this project is to identify and detect 17 different classes distributed across three categories: office, kitchen, and bedroom. During the data collection phase, we obtained a total of 2677 images specifically for the office environment.

The dataset consists of a total of 2677 samples, which are divided into 7 classes representing office environments. The smallest class contains approximately 290 samples. Each class has an adequate number of images distributed across the training, validation, and test sets. The image dataset is organized into three files: train (70\%), validation (20\%), and test (10\%). 

It is important to note that the lack of data can impact the problem of overfitting or underfitting that may arise during the training process. To address these issues, data augmentation techniques are employed. Among the commonly used data augmentation methods, geometric transformations are particularly effective.

\subsection{Transfer Learning}

In the COCO dataset, there are 80 object categories, resulting in an output tensor dimension of 3 x (5 + 80) = 255. Here, 3 represents the number of models for each grid prediction, 5 indicates the coordinates (x, y, w, h), and confidence for each prediction field, and 80 denotes the number of classes in the COCO dataset. In our specific dataset, such as the office environment, we have 7 classes. Hence, the output dimension of the classifier is 3 x (5 + 7).

To address the object detection challenge, we employed YOLOv5 for both detection and classification tasks. We divided the 20 classes into 3 categories, and each category was trained independently using its own model. This approach was adopted to ensure that a relatively small number of classes per category were trained, as the model would be deployed on an embedded system.

We fine-tuned and configured the YOLOv5 architecture specifically for our dataset. To achieve this, we employed transfer learning, adapting the YOLOv5 framework to be compatible with our dataset. We utilized pre-trained weights from a different model that had been trained on the extensive COCO dataset.

For training our model (yolov5s.pt), we utilized the standard Colab VM with 12GB of GPU memory. To enhance the robustness of the trained model and better utilize the available GPU resources, we set the batch size to 4. Additionally, we conducted training for a total of 100 epochs, observing that the trained model reached stability.

Throughout the experiments, we incorporated various hyperparameters. Some of these included weight decay = 0.0005, initial learning rate = 0.0042, final learning rate = 0.1, and momentum = 0.937. These parameters were maintained at their default values. Ultimately, we trained and tested YOLOv5 on the Colab VM using our dataset.

\subsection{Results of objects detection}

\begin{figure}[t]
    \centering
   \includegraphics[width=12cm]{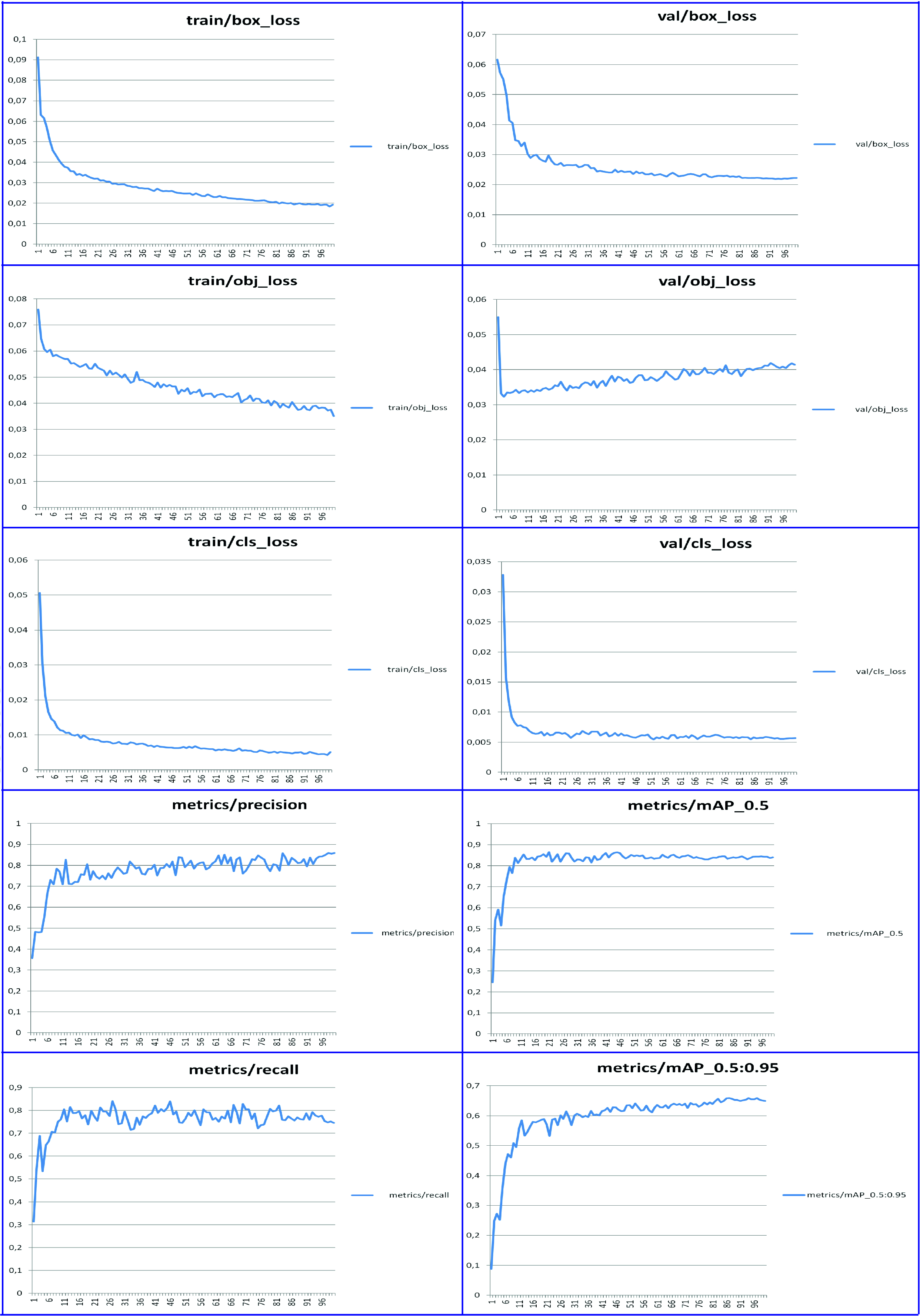}
  \caption{Performance of the fine-tuned YOLOv5 model.}
 \label{fig:result}
\end{figure}

Figure~\ref{fig:result} depicts the performance of the fine-tuned YOLOv5 model during the transfer learning process. The top row of images represents the model's performance on the training set, while the bottom row represents its performance on the validation set.
Upon examining these images, it is evident that the object detection loss for our classes in the training set decreased to a value below 0.02 after 100 epochs. A similar trend can be observed in the validation set, where the object detection loss also decreased over the course of training.

\begin{figure}[t]
  \centering
  \includegraphics[width=10cm]{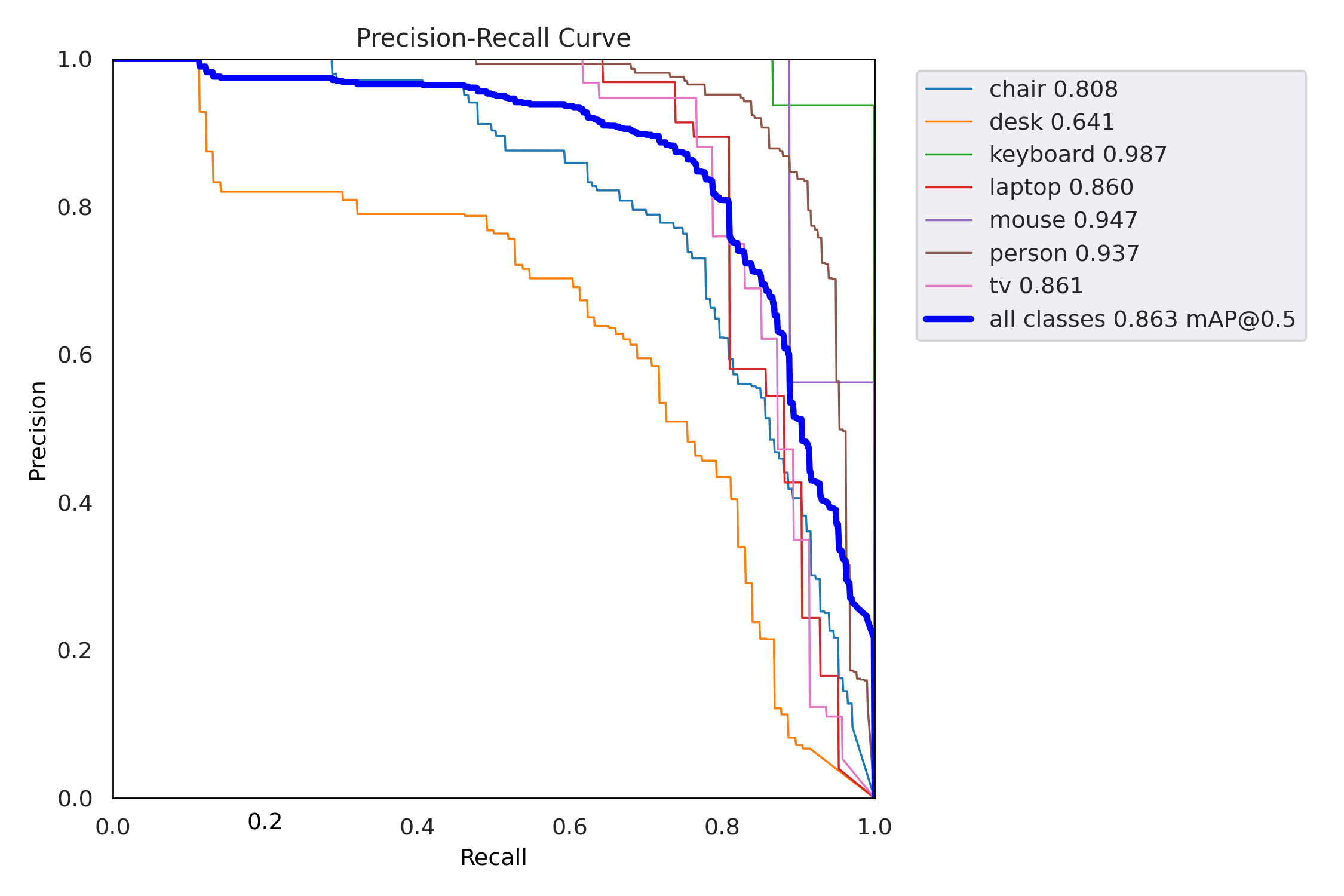}
   \includegraphics[width=10cm]{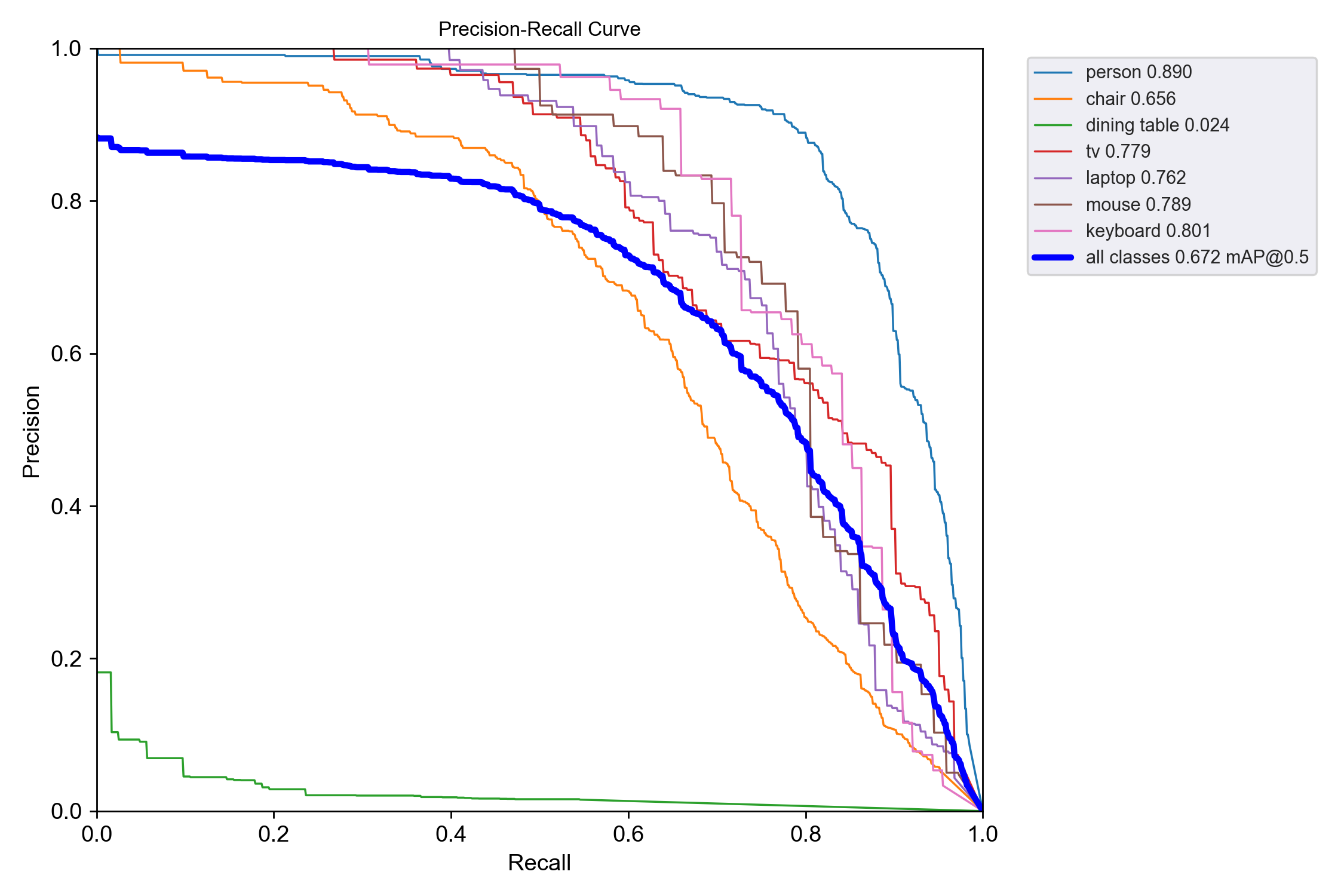}
\caption{Precision, Recall for our model (top) and yolov5s (bottom) on train data.}
  \label{fig:comparaison}
\end{figure}

To provide a more detailed analysis of the model's training process and performance, Figure ~\ref{fig:comparaison}(top) displays a plot showcasing the precision and recall mapping for detecting the seven classes during training. From the figure, it is evident that the model achieved a mean Average Precision (mAP) of 86.3\%. This mAP value represents the area under the curve, indicating the trained model's ability to accurately detect objects with high precision and recall values.

To highlight the superiority of our selected object detection model for the desk environment (comprising the previously mentioned 7 classes), we conducted a comparison with the detection model prior to transfer learning, namely \textbf{yolov5s}. Figure ~\ref{fig:comparaison} showcases the mAP results obtained by both models on a training image set.

It is evident from the results that our model surpasses yolov5s in terms of average precision for the 7 classes. It is important to highlight that the object labeled as \textbf{dining table} in yolov5 is distinct from the \textbf{desk} object. Based on the conducted experiments, our transfer learning model derived from yolov5s exhibits superior performance. Therefore, we can confidently utilize our model for the project.

\subsubsection{Mapping}

The pin grid provides an organized structure and spatial reference for each object based on its position and size. This facilitates further processing or interaction with the detected objects within the project's context.

\begin{figure}[t]
 \centering
  \includegraphics[width=12cm]{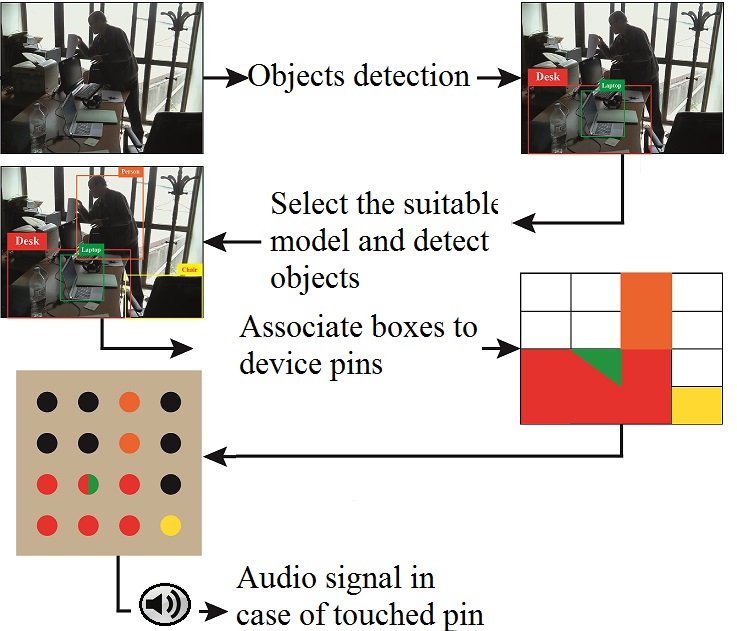}
  \caption{Example of the result of image mapping on the device.}
  \label{fig:mapping}
\end{figure}

Figure ~\ref{fig:mapping} shows how the detected objects in the image  are associated to their corresponding cells. Each object's bounding box is associated to the grid cell as long as the majority of its surface is within that cell. A bounding box can be associated with multiple cells, and likewise, a cell can have multiple bounding boxes .

\begin{figure}[t]
\centering
  \includegraphics[width=10cm]{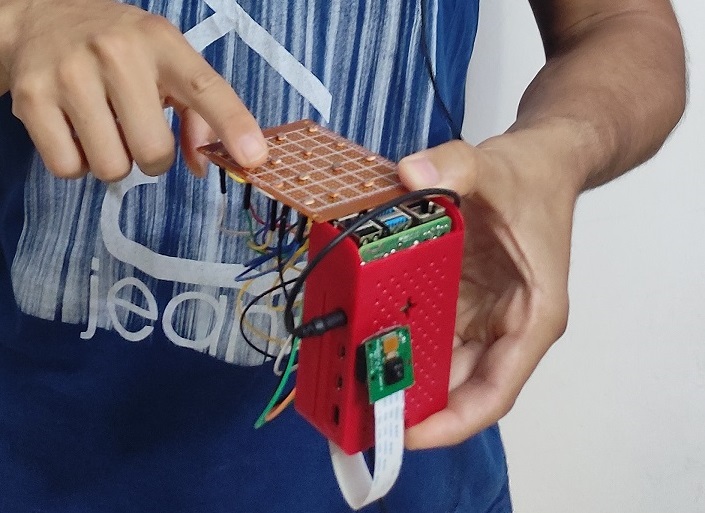}
  \caption{Visually impaired with the tactile device.}
 \label{fig:Un utilisateur malvoyant.}
\end{figure}

\section{Conclusion}

The designed and developed device aims to assist visually impaired individuals by providing information about the objects present in their surroundings, whether they are static or dynamic. Additionally, it enables users to determine their position within the scene.

Tests conducted on the device indicate its usefulness for visually impaired individuals. The system is also capable of capturing images from a TV or desk screen and dynamically mapping the recognized objects in real time on the device.

It's important to note that the current version of the system does not take depth information into account. However, future work will focus on incorporating depth information to accurately locate objects in the device based on their positions in the scene, rather than just as they appear in the image.



\begin{thebibliography}{00}



\bibitem{Ibel2022}
Farah Ibelaiden and Slimane Larabi.
Visual place representation and recognition from depth images.
Optik,260,2022.

\bibitem{Zatout2019}
 Zatout, Chayma and Larabi, Slimane and Mendili, Ilyes and Barnabé, Soedji Ablam Edoh.Ego-Semantic Labeling of Scene from Depth Image for Visually Impaired and Blind People. IEEE/CVF International Conference on Computer Vision Workshop (ICCVW), 2019,
  pp. 4376-4384.
  
\bibitem{Ibel2020}
  Ibelaiden, Farah and Sayah, Brahim and Larabi, Slimane.Scene Description from Depth Images for Visually Positioning.
  2020 1st International Conference on Communications, Control Systems and Signal Processing (CCSSP), 2020, pp.101-106.
  
\bibitem{Benhamida2022}
  Benhamida, Leyla and Larabi, Slimane. Human Action Recognition and Coding based on Skeleton Data for Visually Impaired and Blind People Aid System.
  2022 First International Conference on Computer Communications and Intelligent Systems (I3CIS), 2022, pp. 49-54.
  
\bibitem{Delloul2022_2}
  Delloul, Khadidja and Larabi, Slimane. Egocentric Scene Description for the Blind and Visually Impaired.
  5th International Symposium on Informatics and its Applications (ISIA), 2022. 
  pp. 1-6.
  
\bibitem{Delloul2022}
K. Delloul and S. Larabi. Image Captioning State-of-the-Art: Is It Enough for the Guidance of Visually Impaired in an Environment?
Advances in Computing Systems and Applications, 2022,
pp. 385-394.


\bibitem{Ibel2020_2}
  Ibelaiden, Farah and Larabi, Slimane. A Benchmark for Visual Positioning from Depth Images.
 4th International Symposium on Informatics and its Applications (ISIA), 2020, pp. 1-6.

\bibitem{34}
  Bhole, Swapnil and Dhok, Aniket.
  Deep Learning based Object Detection and Recognition Framework for the Visually-Impaired.
  Fourth International Conference on Computing Methodologies and Communication (ICCMC), 2020, pp. 725-728.
  
\bibitem{31}
Hegde, Pavan, et al.
Smart Glasses for Visually Disabled Person.
Journal of Research in Engineering and Science (IJRES), 9(7), pp. 62-68, 2021.

\bibitem{benhamida2023theater}
Leyla Benhamida and Slimane Larabi. Theater Aid System for the Visually Impaired Through Transfer Learning of Spatio-Temporal Graph Convolution Networks. 
arXiv 2023, eprint. 2306.16357.

\bibitem{CCSSP2020}
Zatout, Chayma and Larabi, Slimane.
A Novel Output Device for visually impaired and blind people’s aid systems. 
2020 1st International Conference on Communications, Control Systems and Signal Processing (CCSSP), pp. 119-124.

\bibitem{these-chaymal-arabi}
Zatout, Chayma and Larabi, Slimane.
Semantic scene synthesis: application to assistive systems.
The Visual Computer, 38, pp. 2691-2705, 2022.


\bibitem{Arkin2022}
  Ershat Arkin, Nurbiya Yadikar, Xuebin Xu, Alimjan Aysa and Kurban Ubul.
  Object detection methods from CNN to transformer. 
  Multimedia Tools and Applications, 1, 2022,  pp.  1573-7721.

\bibitem{Jiao_2019}
Licheng Jiao and Fan Zhang and Fang Liu and Shuyuan Yang and Lingling Li and Zhixi Feng and Rong Qu. A Survey of Deep Learning Based Object Detection.
IEEE Access, 2019,(7), pp. 128837-128868.

\bibitem{girshick2014rich}
Girshick, Ross and Donahue, Jeff and Darrell, Trevor and Malik, Jitendra.
Rich Feature Hierarchies for Accurate Object Detection and Semantic Segmentation. 2014 IEEE Conference on Computer Vision and Pattern Recognition, pp. 580-587.

\bibitem{girshick2015fast}
Girshick, Ross. Fast R-CNN. 
  2015 IEEE International Conference on Computer Vision (ICCV), pp. 1440-1448.

\bibitem{LARABI2009}
Slimane Larabi. Textual description of shapes.
Journal of Visual Communication and Image Representation, 20(8), pp. 563-584,
2009.

\bibitem{ren2016faster}
 Ren, Shaoqing and He, Kaiming and Girshick, Ross and Sun, Jian.
 Faster R-CNN: Towards Real-Time Object Detection with Region Proposal Networks.
 Advances in Neural Information Processing Systems, 28, 2015.


\bibitem{8417976}
Lin, Tsung-Yi and Goyal, Priya and Girshick, Ross and He, Kaiming and Dollár, Piotr.
  Focal Loss for Dense Object Detection.
 IEEE Transactions on Pattern Analysis and Machine Intelligence,42(2),pp. 318-327, 2020.

\bibitem{Liu_2016}
	Wei Liu and Dragomir Anguelov and Dumitru Erhan and Christian Szegedy and Scott Reed and Cheng-Yang Fu and Alexander C. Berg. SSD: Single Shot {MultiBox} Detector. Computer Vision ECCV, 2016, pp. 21-37
 
\bibitem{wang2021scaledyolov4}
Chien-Yao Wang and Alexey Bochkovskiy and Hong-Yuan Mark Liao.
Scaled-YOLOv4: Scaling Cross Stage Partial Network. 
arXiv, eprint. 2011.08036, 2021.


\bibitem{dosovitskiy2021image}
Alexey Dosovitskiy et al. 
An Image is Worth 16x16 Words: Transformers for Image Recognition at Scale.
arXiv, eprint. 2010.11929, 2021.


\bibitem{Lin2014}
Lin, Tsung-Yi et al.",
Microsoft COCO: Common Objects in Context.
Computer Vision ECCV 2014, pp. 740-755.


\bibitem{naftali2022}
Martinus Grady Naftali and Jason Sebastian Sulistyawan and Kelvin Julian.
Comparison of Object Detection Algorithms for Street-level Objects. 
arXiv, eprint. 2208.11315, 2022.

\bibitem{Banf2016}
Michael Banf, Ruben Mikalay, Baris Watzke and Volker Blanz.,
PictureSensation – a mobile application to help the blind explore the visual
world through touch and sound. Journal of Rehabilitation and Assistive
Technologies Engineering, (3), 2016, pp. 1–10.

\bibitem{Diwan2023}
Tausif Diwan and Grandhi Sai Anirudh and Jitendra V. Tembhurne.
Object detection using YOLO: challenges, architectural successors, datasets and applications. Multimedia Tools and Applications, 82, pp. 9243 - 9275, 2023.

\bibitem{CAVACO2013}
Sofia Cavaco and J. Tomás Henriques and Michele Mengucci and Nuno Correia and Francisco Medeiros. 
Color Sonification for the Visually Impaired.
Procedia Technology,(9), pp. 1048-1057,2013.

\bibitem{Banf2013}
Banf, Michael and Blanz, Volker.
Sonification of Images for the Visually Impaired Using a Multi-Level Approach.
Proceedings of the 4th Augmented Human International Conference, 2013, pp. 162–169.


\end{thebibliography}
\end{document}